\documentclass[conference]{IEEEtran}
\IEEEoverridecommandlockouts
\usepackage{cite}
\usepackage{amsmath,amssymb,amsfonts}
\usepackage{algorithmic}
\usepackage{booktabs}
\usepackage{graphicx}
\usepackage{textcomp}
\usepackage{xcolor}
\usepackage{subcaption}
\usepackage{graphicx}
\usepackage{tablefootnote}
\usepackage{hyperref}
\usepackage{multirow}
\usepackage{array}
\usepackage{subcaption}
\usepackage{multicol}

\def\BibTeX{{\rm B\kern-.05em{\sc i\kern-.025em b}\kern-.08em
    T\kern-.1667em\lower.7ex\hbox{E}\kern-.125emX}}
\begin{document}

\title{Boosting the Performance of Transformer Architectures for Semantic Textual Similarity}

\author{
\IEEEauthorblockN{Ivan Rep}
\IEEEauthorblockA{\textit{Department of Electronics, Microelectronics}, \\ \textit{Computer and Intelligent Systems} \\
\textit{Faculty of Electrical Engineering and Computing}\\
Zagreb, Croatia\\
\texttt{ivan.rep2@fer.hr}}
\and
\IEEEauthorblockN{Assoc. prof. Vladimir Čeperić}
\IEEEauthorblockA{\textit{Department of Electronics, Microelectronics}, \\ \textit{Computer and Intelligent Systems} \\
\textit{Faculty of Electrical Engineering and Computing}\\
Zagreb, Croatia\\
\texttt{vladimir.ceperic@fer.hr}}
}

\maketitle

\begin{abstract}
Semantic textual similarity is the task of estimating the similarity between the meaning of two texts. In this paper, we fine-tune transformer architectures for semantic textual similarity on the Semantic Textual Similarity Benchmark by tuning the model partially and then end-to-end. We experiment with BERT, RoBERTa, and DeBERTaV3 cross-encoders by approaching the problem as a binary classification task or a regression task. We combine the outputs of the transformer models and use handmade features as inputs for boosting algorithms. Due to worse test set results coupled with improvements on the validation set, we experiment with different dataset splits to further investigate this occurrence. We also provide an error analysis, focused on the edges of the prediction range.
\end{abstract}


\section{Introduction}
In any natural language processing task, it is difficult to encapsulate understanding text into a series of clear steps, which is why many approaches exploit machine learning. One of the tasks in natural language processing is determining the semantic similarity of texts in which the system estimates how similar the texts are by their meaning. A common technique coupled with these tasks is the usage of transformer architectures. These models exploit the attention mechanism by estimating the connection of each word to a different word in the text. Combining these architectures and representing words as embeddings proved to be a powerful tool. These models are usually pretrained on large text corpora using different pretraining techniques.\\

\indent In this paper we explore fine-tuning transformer architectures on the \emph{Semantic Textual Similarity Benchmark} (STSB) dataset \cite{STSB} \cite{STSB2}. Our approach consists of employing different architectures and different sizes of these architectures. What we do differently is that we combine these models with gradient boosting algorithms which are state-of-the-art algorithms for learning from tabular data. Besides using outputs of multiple models as inputs for the gradient boosting algorithm, we also use handcrafted features that model structure instead of semantics. Another deviation from the usual approach is that we train the transformers by only fine-tuning the regression head, followed by end-to-end fine-tuning. The rest of this paper is organized as follows: Section 2 presents the related work. In Section 3. we describe our experimental setup. Section 4. presents the results achieved by transformer architectures and boosting algorithms. In Section 5. we discuss and analyse the results and explore different data splits to get more consistent results. Finally, in Section 6. we present our conclusions and suggest future improvements.


\setlength\extrarowheight{1pt}
\begin{table*}[h]
  \centering 
\caption{Hyperparameters chosen for the final models using mean squared error}
\begin{tabular}{l|ccc|ccc}
     & \multicolumn{3}{c}{\textbf{Partial fine-tuning}} & \multicolumn{3}{c}{\textbf{End-to-end fine-tuning}}\\
    \hline
    \textbf{Model} &  \textbf{Batch size} & \textbf{Learning rate} & \textbf{Weight decay} & \textbf{Batch size} & \textbf{Learning rate} & \textbf{Weight decay}\\
    \hline
    BERT base cased        & 32  & 5e-4  & 1e-4  & 32  & 5e-5  & 1e-4 \\
    BERT large cased       &  8 & 5e-4 & 1e-2   & 16   & 1e-5  & 1e-3 \\
    RoBERTa base            &  32 & 5e-4 & 1e-4 & 32 & 5e-5 & 1e-4 \\
    RoBERTa large           &  32 & 5e-4 & 1e-4 & 32 & 5e-5 & 1e-4 \\
    DistilBERT base cased & 32  & 5e-4 & 1e-4 & 32 & 5e-5& 1e-4 \\
    DistilRoBERTa base     &32& 5e-4&1e-4 & 32& 5e-5 & 1e-4  \\
    DeBERTaV3 small        & 32& 5e-4 & 1e-4   &8 &5e-5& 1e-2 \\
    DeBERTaV3 base        & 32& 5e-4& 1e-4   &   8 &  5e-5&  1e-3 \\
    DeBERTaV3 large       &8 & 5e-4& 1e-2   & 8 & 1e-5 &1e-4\\
    \hline
\end{tabular}
\label{mse-hyperparams}
\end{table*}

\setlength\extrarowheight{1pt}
\begin{table*}[h]
  \centering 
\caption{Hyperparameters chosen for the final models using cross-entropy error}
\begin{tabular}{l|ccc|ccc}
     & \multicolumn{3}{c}{\textbf{Partial fine-tuning}} & \multicolumn{3}{c}{\textbf{End-to-end fine-tuning}}\\
    \hline
    \textbf{Model} &  \textbf{Batch size} & \textbf{Learning rate} & \textbf{Weight decay} & \textbf{Batch size} & \textbf{Learning rate} & \textbf{Weight decay}\\
    \hline
    BERT base cased        &  8 & 5e-4   &1e-2  &  8              & 5e-5             & 1e-2 \\
    BERT large cased       &   32 & 5e-4& 1e-2 & 8              & 1e-5              &1e-4 \\
    RoBERTa base            &  32 & 5e-4& 1e-4 &  8              & 5e-5              & 1e-2  \\
    RoBERTa large           &   32& 5e-4& 1e-4 &  8              & 1e-5              & 1e-2 \\
    DistilBERT base cased & 8  & 5e-4 & 1e-2 & 8              & 5e-5             & 1e-2   \\
    DistilRoBERTa base     & 32  &5e-4&1e-4 &   8              & 5e-5              &1e-2 \\
    DeBERTaV3 small        & 32 & 5e-4 &1e-4 &   28             & 5e-5            & 1e-4 \\
    DeBERTaV3 base        & 32 & 5e-4& 1e-4& 32             & 5e-5              & 1e-4\\
    DeBERTaV3 large       & 32 &5e-4 & 1e-4 &8              & 1e-5              & 1e-3 \\
    \hline
\end{tabular}
\label{ce-hyperparams}
\end{table*}


\section{Related Work}
With the advancement of representation learning techniques like BERT \cite{BERT}, a lot of natural language processing has shifted towards deep learning. This approach combines masked language modeling and next-sentence prediction to pretrain a transformer architecture in an unsupervised fashion. The RoBERTa \cite{ROBERTA} model does a replication study with smaller modifications: removing the next sentence prediction objective and training the model longer, with bigger batches and more data. DistilBERT \cite{DISTILBERT} is a different approach that considers the size of BERT models. It uses smaller models and knowledge distillation to retain most of the performances while being faster. DeBERTa \cite{DEBERTA} is a technique that uses two novelties: the disentangled attention mechanism, where words are represented using vectors for content encoding and position encoding, and the modified mask decoder to incorporate the absolute positions in the decoding layer used in pretraining. DeBERTaV3 \cite{DEBERTAV3}, an improvement of the original model, replaces masked language modeling with replaced token detection. The most computationally efficient approach for this task is the SentenceBERT\cite{SENTENCEBERT} architecture. This model derives sentence embeddings which can be compared using cosine similarity. This method is more efficient because the calculated embeddings are independent of one another before calculating the cosine similarity. Each architecture was also fine-tuned for semantic textual similarity, while some achieved state-of-the-art results. Before these advancements, some of the used methods were feature-based \cite{EVOLUTION}. The features used for these systems usually relied on lexical resources such as WordNet \cite{WORDNET} and were heavily dependent on ontologies with semantic features. Most of these methods can be combined with ensembles. Similar approaches have been a research topic in different areas of deep learning \cite{ENSEMBLING}. \\

On the other hand, boosting algorithms are an older approach to learning from tabular data that has recently seen a rise in popularity due to the development of new algorithms. The AdaBoost approach \cite{ADABOOST} combines weak learners where subsequent weak learners are tweaked to mitigate the error of previous classifiers and offers resistance to overfitting. One of the more recent approaches is XGBoost \cite{XGBOOST}, which uses sparsity awareness and weighted quantile sketch for approximate tree learning. Finally, LightGBM \cite{LGBM} offers a different approach to boosting algorithms by incorporating two novel techniques: gradient-based one-side sampling (GOSS) and exclusive feature bundling (EFB). GOSS uses the exclusion of data with small gradients, while EFB bundles together mutually exclusive features to select a smaller number of features.\\

\setlength\extrarowheight{1pt}
\begin{table}[b]
\begin{center}
\caption{Dataset breakdown according to the original datasets and years}
\begin{tabular}{ccc|ccc}
\textbf{Genre} & \textbf{File} & \textbf{Years}& \textbf{Train} & \textbf{Dev} & \textbf{Test} \\
\hline
 news   &  MSRpar      &  2012&    1000& 250& 250\\
 news     & headlines     & 2013-16 & 1999  &250&  250 \\
 news     & deft-news     & 2014     & 300 &   0   & 0\\
 captions & MSRvid         &2012    & 1000 & 250 & 250\\
 captions  &images        & 2014-15  &1000 & 250 & 250\\
 captions  &track5.en-en  & 2017     &   0 & 125  &125\\
 forum    & deft-forum    & 2014     & 450   & 0   & 0\\
 forum    & answers-forums &2015   &     0 & 375 &   0\\
 forum    & answer-answer & 2016     &   0&    0  &254\\
\hline
\end{tabular}
\label{stsb-table}
\end{center}
\end{table}

\section{Experimental Setup}
\subsection{Dataset and Metrics}
As previously mentioned, the dataset used for all the experiments is the \emph{Semantic Textual Similarity Benchmark} dataset. This dataset is a collection of English datasets used for semantic textual similarity tasks in SemEval between 2012 and 2017. The texts are from various sources like image captions, news headlines, and user forums. The dataset contains 5749 examples in the train split, 1500 in the dev split, and 1379 in the test splits, summing up to 8628 sentence pairs. Since this task is a regression task, the given labels are in the [0, 5] range. A more detailed examination of the dataset is shown in table \ref{stsb-table}. An interesting observation is that some datasets only appear in some dataset splits.\\

The metrics used for this task are Pearson's correlation coefficient (denoted by \ref{pearson}), which measures the linear correlation between the variables, and Spearman's rank correlation coefficient (denoted by \ref{spearman}), which represents the Pearson's correlation coefficient applied to ranks of observations.

\begin{equation}
r_p =  \frac{\sum_{i=1}^{n} (x_i - \overline{x}) (y_i - \overline{y})}{\sum_{i=1}^{n}( x_i - \overline{x}) \sum_{i=1}^{n} (y_i - \overline{y})}
\label{pearson}
\end{equation}

\begin{equation}
r_s = 1 - \frac{6}{n(n^2 - 1)} \sum_{i=1}^{n} d_i^2
\label{spearman}
\end{equation}

In equation \eqref{pearson}, $x_i$ and $y_i$ represent observation $i$, $\overline x$ and $\overline y$ represent the means of corresponding variables., while in equation \eqref{spearman}, $n$ represents the number of observations and $d_i$ represents the rank difference of observations $x_i$ and $y_i$.

\setlength\extrarowheight{1pt}
\begin{table*}[h]
  \centering 
\caption{Transformer results using using mean squared error}
\begin{tabular}{l|ccc|ccc}
     & \multicolumn{3}{c}{\textbf{Partial fine-tuning}} & \multicolumn{3}{c}{\textbf{End-to-end fine-tuning}}\\
    \hline
    \textbf{Model} & \textbf{Train set} & \textbf{Dev set} & \textbf{Test set} & \textbf{Train set} & \textbf{Dev set} & \textbf{Test set}\\
    \hline
    BERT base cased        & 0.793/0.749 & 0.814/0.809 & 0.735/0.697 & 0.995/0.995 & 0.899/0.896 & 0.865/0.856\\
    BERT large cased       & 0.790/0.754 & \textbf{0.824}/\textbf{0.823} & 0.731/0.694 & 0.993/0.992 & 0.908/0.904 & 0.869/0.857\\
    RoBERTa base            & 0.631/0.629 & 0.585/0.591 & 0.569/0.578 & 0.989/0.988 & 0.913/0.911 & 0.895/0.890\\
    RoBERTa large           & 0.512/0.506  & 0.492/0.486 & 0.493/0.502 & 0.994/0.994 & 0.921/0.920 & 0.904/0.899\\
    DistilBERT base cased & 0.639/0.604  & 0.604/0.605 & 0.579/0.555 & 0.994/0.993 & 0.863/0.861 & 0.814/0.800\\
    DistilRoBERTa base     & 0.576/0.581  & 0.485/0.477 & 0.510/0.516 & 0.988/0.986 & 0.887/0.885 & 0.858/0.849\\
    DeBERTaV3 small        & 0.782/0.764 & 0.761/0.763 & 0.758/0.758 & 0.991/0.990 & 0.906/0.904 & 0.892/0.888 \\
    DeBERTaV3 base         & \textbf{0.838}/\textbf{0.832} & 0.809/\textbf{0.823} & \textbf{0.824}/\textbf{0.83}1 & \textbf{0.996}/\textbf{0.996} & 0.917/0.915 & 0.907/0.904\\ 
    DeBERTaV3 large        & 0.828/0.822  & 0.807/0.816 & 0.820/0.825 & 0.991/0.990 & \textbf{0.927}/\textbf{0.926} & \textbf{0.922}/\textbf{0.921}\\ 
    \hline
\end{tabular}
\label{mse-results}
\end{table*}

\setlength\extrarowheight{1pt}
\begin{table*}[h]
  \centering 
\caption{Transformer results using using cross-entropy error}
\begin{tabular}{l|ccc|ccc}
     & \multicolumn{3}{c}{\textbf{Partial fine-tuning}} & \multicolumn{3}{c}{\textbf{End-to-end fine-tuning}}\\
    \hline
    \textbf{Model} & \textbf{Train set} & \textbf{Dev set} & \textbf{Test set} & \textbf{Train set} & \textbf{Dev set} & \textbf{Test set}\\
    \hline
    BERT base cased        & 0.779/0.723 & 0.813/0.802 & 0.726/0.686 & \textbf{0.997}/\textbf{0.996} & 0.899/0.896 & 0.861/0.849 \\
    BERT large cased       & 0.791/0.751 & \textbf{0.822}/\textbf{0.821}  & 0.736/0.699 & 0.978/0.976 & 0.909/0.906 & 0.875/0.865 \\
    RoBERTa base            & 0.621/0.612 &  0.575/0.577 & 0.567/0.572 &  0.992/0.991 & 0.908/0.906 & 0.886/0.881\\
    RoBERTa large           & 0.513/0.520 &  0.484/0.478 & 0.494/0.513 & 0.989/0.988 & 0.924/0.923 & 0.913/0.909 \\
    DistilBERT base cased &  0.656/0.614 & 0.609/0.614 & 0.589/0.561 & 0.991/0.990 & 0.860/0.856 & 0.814/0.801 \\
    DistilRoBERTa base     &  0.590/0.586  & 0.488/0.475  & 0.515/0.510 & 0.990/0.989 & 0.890/0.888 & 0.859/0.850  \\
    DeBERTaV3 small        & 0.793/0.767 & 0.769/0.767 & 0.770/0.762 & 0.990/0.988  & 0.907/0.904 & 0.893/0.890 \\
    DeBERTaV3 base         &  \textbf{0.843}/\textbf{0.830}  & 0.813/0.819 & \textbf{0.828}/\textbf{0.827} & 0.988/0.987 & 0.919/0.917 & 0.912/0.911 \\ 
    DeBERTaV3 large        & 0.835/0.821 & 0.814/0.817 & 0.826/0.824 & 0.986/0.984 & \textbf{0.927}/\textbf{0.926} & \textbf{0.919}/\textbf{0.919} \\ 
    \hline
\end{tabular}
\label{ce-results}
\end{table*}

\subsection{Model Fine-Tuning}
The fine-tuning approach we took is fine-tuning with the frozen base model followed by end-to-end fine-tuning. The regression head weights are randomly initialized, so the idea was to make the model more used to the data it is going to be fine-tuned on. This procedure is followed by end-to-end fine-tuning that tunes all the parameters of the transformer architecture. \\

The hyperparameter optimization for fine-tuning with the frozen base model was done using exclusively Population-based training \cite{PBT}, while end-to-end fine-tuning was done using a combination of Population-based training and hand tuning. The Population-based training method was preferred, but due to hardware constraints, the RoBERTa large, BERT large cased, DeBERTaV3 base, and DeBERTaV3 large were tuned by hand. Final selected hyperparameters can be seen in tables \ref{mse-hyperparams} and \ref{ce-hyperparams}.\\

Two approaches were explored:
\begin{itemize}
\item Framing the problem as a regression problem by using mean squared error
\item Framing the problem as a binary classification problem by using a cross-entropy error
\end{itemize}
The second approach also required scaling the labels to range [0, 1].\\

Each model was fine-tuned for ten epochs with the addition of early stopping if no improvement occurs in the last three epochs. The scheduler used was the cosine annealing scheduler with a 10\% warmup ratio. FP16 training was used to reduce computation time and help with the hardware constraints for all the models. Also, every model was optimized using Adam with weight decay regularization \cite{ADAMW}.

\setlength\extrarowheight{1pt}
\begin{table}[b]
\begin{center}
\caption{Results using the baselines}
\begin{tabular}{ccccc}
\textbf{Model} & \textbf{Train set} & \textbf{Dev set} & \textbf{Test set} \\
\hline
 Cosine similarity & 0.459/0.462 & \textbf{0.478}/\textbf{0.540} & \textbf{0.367}/\textbf{0.388} \\
 Linear regression  & 0.440/0.425 & 0.119/0.118 & 0.194/0.193 \\
 Support vector machine & \textbf{0.585}/\textbf{0.576} & 0.258/0.240 & 0.330/0.301 \\
\end{tabular}
\label{baseline-results}
\end{center}
\end{table}

\begin{table*}[t]
  \centering 
\caption{Boosting results using using mean squared error transformers}

\setlength\tabcolsep{4.3pt}
\begin{tabular}{l|cccc|cccc}
     & \multicolumn{4}{c}{\textbf{Using 2 transformers}} & \multicolumn{4}{c}{\textbf{Using 3 transformers}}\\
    \hline
    \textbf{Model} & \textbf{Train set} & \textbf{Dev set} & \textbf{Test set} & \textbf{Transformers used} & \textbf{Train set} & \textbf{Dev set} & \textbf{Test set} & \textbf{Transformers used}\\
    \hline
    AdaBoost & 0.993/0.995 & 0.927/0.926 & 0.911/0.908 & DeBERTaV3, RoBERTa &0.994/0.996 & 0.927/0.925 & 0.904/0.899 & BERT, DeBERTaV3, RoBERTa \\
    XGBoost &  \textbf{0.996}/\textbf{0.996} & \textbf{0.929}/\textbf{0.927} & 0.912/0.909 & DeBERTaV3, RoBERTa & \textbf{0.997}/\textbf{0.997} & 0.929/0.927 & \textbf{0.910}/\textbf{0.906} & BERT, DeBERTaV3, RoBERTa\\
    LightGBM & \textbf{0.996}/\textbf{0.996} & \textbf{0.929}/\textbf{0.927} & \textbf{0.915}/\textbf{0.913} & DeBERTaV3, RoBERTa & \textbf{0.997}/\textbf{0.997} & \textbf{0.930}/\textbf{0.928} & \textbf{0.910}/\textbf{0.906} & BERT, DeBERTaV3, RoBERTa \\
    \hline
\end{tabular}
\label{boost-mse-results}
\end{table*}

\begin{table*}[t]
  \centering 
\caption{Boosting results using using cross-entropy error transformers}
\setlength\tabcolsep{5pt}
\begin{tabular}{l|cccc|cccc}
     & \multicolumn{4}{c}{\textbf{Using 2 transformers}} & \multicolumn{4}{c}{\textbf{Using 3 transformers}}\\
    \hline
    \textbf{Model} & \textbf{Train set} & \textbf{Dev set} & \textbf{Test set} & \textbf{Transformers used} & \textbf{Train set} & \textbf{Dev set} & \textbf{Test set} & \textbf{Transformers used}\\
    \hline
    AdaBoost & 0.988/0.989 & 0.929/0.929 & 0.915/0.913 & BERT, DeBERTaV3 & 0.992/0.993 & 0.930/0.930 & 0.919/0.917 & BERT, DeBERTaV3, RoBERTa \\
    XGBoost & \textbf{0.991}/\textbf{0.990} & 0.931/0.929 & 0.914/0.910 & BERT, DeBERTaV3 & \textbf{0.994}/\textbf{0.993} & 0.932/\textbf{0.931} & 0.919/0.915 & BERT, DeBERTaV3, RoBERTa\\
    LightGBM & 0.990/0.989 & \textbf{0.933}/\textbf{0.931} & \textbf{0.918}/\textbf{0.916} & BERT, DeBERTaV3 & \textbf{0.994}/\textbf{0.993} & \textbf{0.933}/\textbf{0.931} & \textbf{0.921}/\textbf{0.918} & BERT, DeBERTaV3, RoBERTa\\
    \hline
\end{tabular}
\label{boost-ce-results}
\end{table*}


\begin{table*}[h]
  \centering 
\caption{Results using cross-entropy error transformers with a stratified cross-validation split}

\begin{tabular}{l|ccc|ccc}
     & \multicolumn{3}{c}{\textbf{Partial fine-tuning}} & \multicolumn{3}{c}{\textbf{End-to-end fine-tuning}}\\
    \hline
    \textbf{Model} & \textbf{Train set} & \textbf{Dev set} & \textbf{Test set} & \textbf{Train set} & \textbf{Dev set} & \textbf{Test set}\\
    \hline
    BERT large cased & 0.798/0.768 & 0.793/0.761 & 0.794/0.760 & \textbf{0.994}/\textbf{0.994} & 0.908/0.898 & 0.904/0.890\\
    RoBERTa large & 0.550/0.563 & 0.526/0.541 & 0.534/0.534 & 0.984/0.984 & 0.932/0.927 & 0.926/0.919\\
    DeBERTaV3 large & \textbf{0.829}/\textbf{0.821} & \textbf{0.834}/\textbf{0.831} & \textbf{0.833}/\textbf{0.824} & \textbf{0.994}/0.993 & \textbf{0.937}/\textbf{0.933} & \textbf{0.936}/\textbf{0.928}\\
\end{tabular}
\label{stratified-ce-results}
\end{table*}

\begin{table*}[h]
  \centering 
\caption{Boosting results using using cross-entropy error transformers with a stratified cross-validation split}
\setlength\tabcolsep{5pt}
\begin{tabular}{l|cccc|cccc}
     & \multicolumn{4}{c}{\textbf{Using 2 transformers}} & \multicolumn{4}{c}{\textbf{Using 3 transformers}}\\
    \hline
    \textbf{Model} & \textbf{Train set} & \textbf{Dev set} & \textbf{Test set} & \textbf{Transformers used} & \textbf{Train set} & \textbf{Dev set} & \textbf{Test set} & \textbf{Transformers used}\\
    \hline
    AdaBoost & 0.995/\textbf{0.996} & 0.922/0.917 & 0.918/0.909 & BERT, DeBERTaV3 & 0.995/0.996 & 0.925/0.920 & 0.921/0.912 & BERT, DeBERTaV3, RoBERTa\\
    XGBoost & \textbf{0.997}/\textbf{0.996} & 0.931/0.926 & 0.926/0.917 & BERT, DeBERTaV3 & \textbf{0.997}/\textbf{0.997} & 0.931/0.927 & 0.926/0.917 & BERT, DeBERTaV3, RoBERTa\\
    LightGBM & \textbf{0.997}/\textbf{0.996} & \textbf{0.933}/\textbf{0.927} & \textbf{0.930}/\textbf{0.920} & BERT, DeBERTaV3 & \textbf{0.997}/0.996 & \textbf{0.934}/\textbf{0.928} & \textbf{0.930}/\textbf{0.921} & BERT, DeBERTaV3, RoBERTa\\
    \hline
\end{tabular}
\label{boost-stratified-ce-results}
\end{table*}

\subsection{Boosting}
After training the transformer models, their outputs were used as features combined with handcrafted features. The handcrafted features chosen were focused on the structure of sentences because transformers encode semantics very well. The chosen features were counts of: 
\begin{itemize}
\item characters
\item stopwords
\item tokens
\item verbs
\item adjectives
\item overlapping tokens
\item overlapping lemmas\\
\end{itemize}

The first two features were chosen due to their simplicity and to provide some basic information about the length and composition of the sentences. The reason for counts of tokens is that transformer architectures are only aware of the number of subwords due to the subword tokenization algorithm, and the token count provides a way to capture some information about the original words in the sentence. The motivation for counts of verbs is similar to the previous reason, but also because the models do not have explicit knowledge of POS tags, and therefore counting verbs can provide more detail about the syntax of a sentence. Finally, overlapping tokens and lemmas were included to capture some of the semantic and syntactic information in the text, especially when similar sentences use different words with similar meanings. This information might help the boosting algorithm to understand the sentences better and make more accurate predictions. \\

The boosting algorithms chosen for the task were AdaBoost, XGBoost, and LightGBM. Hyperparameter optimization was carried out, although using grid search. Only combinations of two and three transformers were considered.


\begin{figure*}[t]
\caption{The approximated densities for the Jaccard index of lemmas, the Jaccard index of lemmas without stopwords, percentage of meaningful lemmas in the first sentence, percentage of meaningful lemmas in the second sentence. The blue curve represents the correct predictions, while the orange curve represents the incorrect predictions. The plots in the first row use the best model trained with mean squared error, while the plots in the second row use the best one with cross-entropy error.}
\label{kdeplots}
\begin{center}
\begin{multicols}{4}

    \includegraphics[width=1\linewidth]{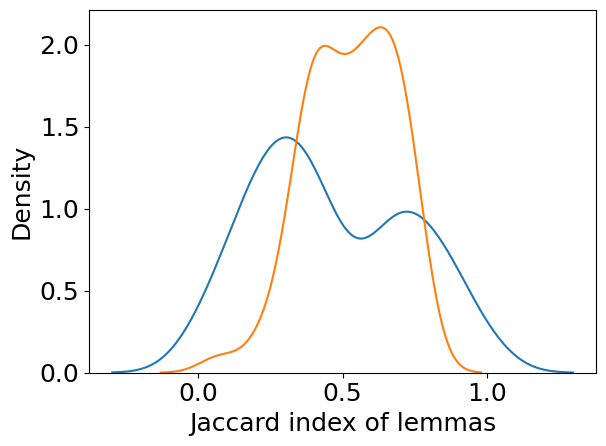}\par 
    \includegraphics[width=1\linewidth]{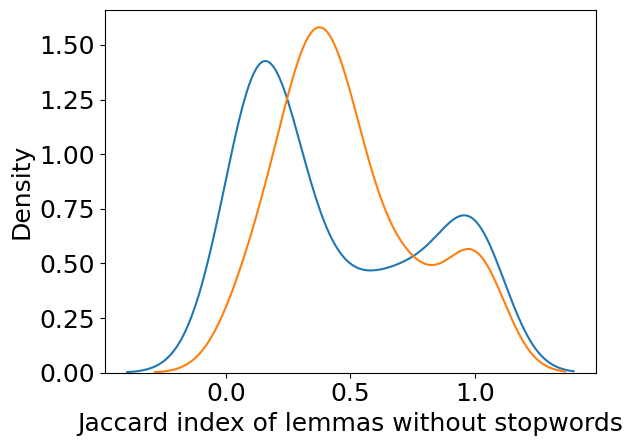}\par     
    \includegraphics[width=1\linewidth]{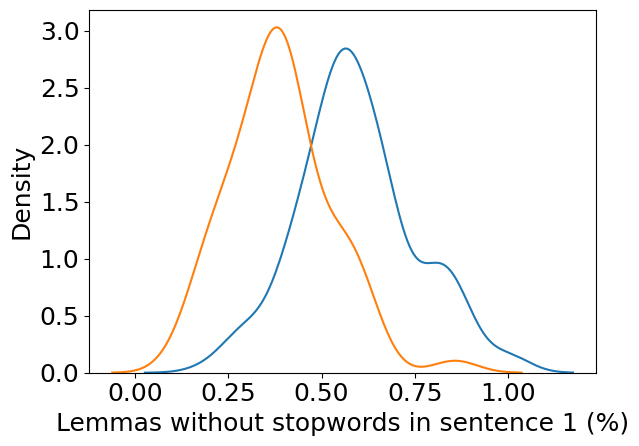}\par
    \includegraphics[width=1\linewidth]{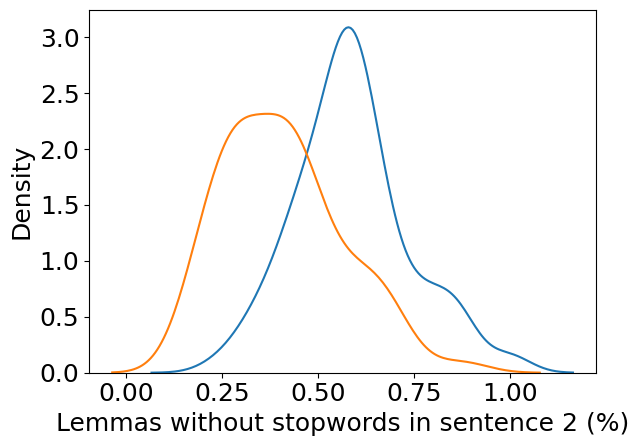}\par 
\end{multicols}
\begin{multicols}{4}
    \includegraphics[width=1\linewidth]{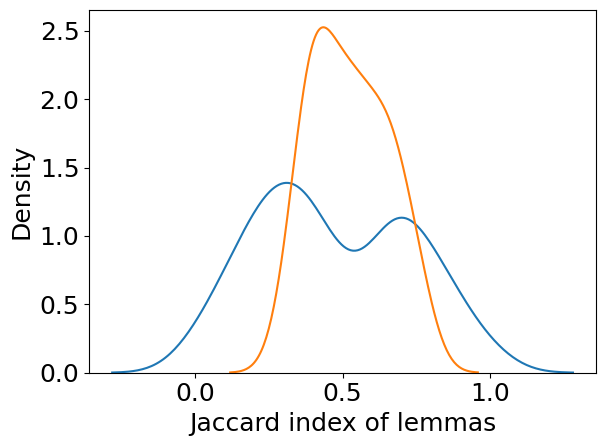}\par 
    \includegraphics[width=1\linewidth]{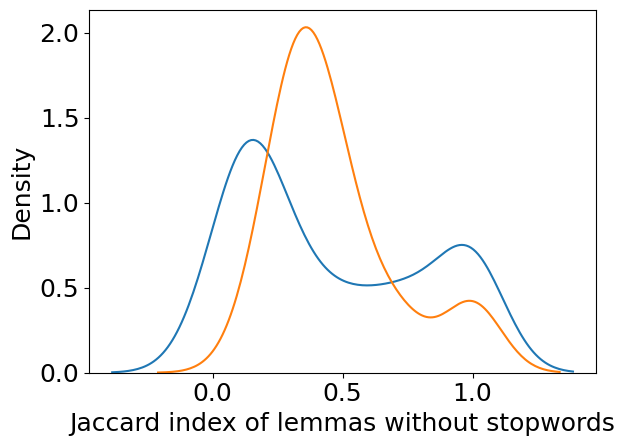}\par     
    \includegraphics[width=1\linewidth]{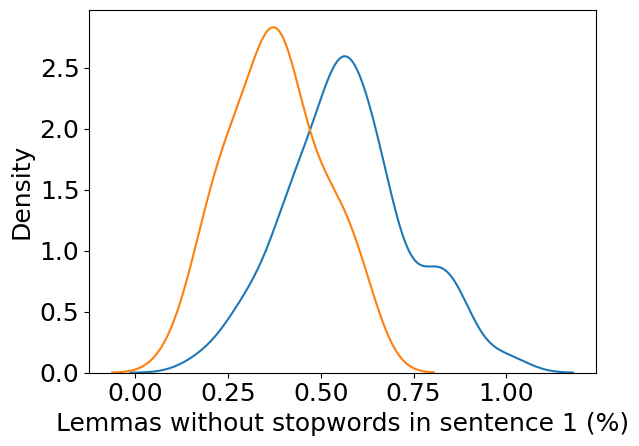}\par
    \includegraphics[width=1\linewidth]{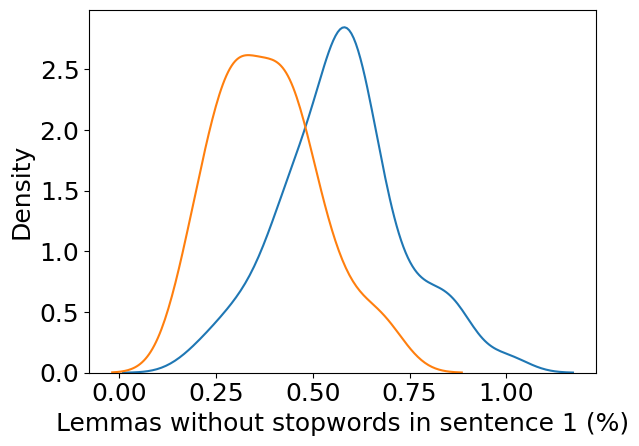}\par 
\end{multicols}
\end{center}
\end{figure*}

\setlength{\textfloatsep}{0pt}
\begin{figure}[t]
\caption{Scatter plot of predictions and labels. The left plot represents the mean squared error model, while the right plot represents the cross-entropy error model.}
\label{fig:scatterplots}
\begin{center}
\begin{multicols}{2}
    \includegraphics[width=1\linewidth]{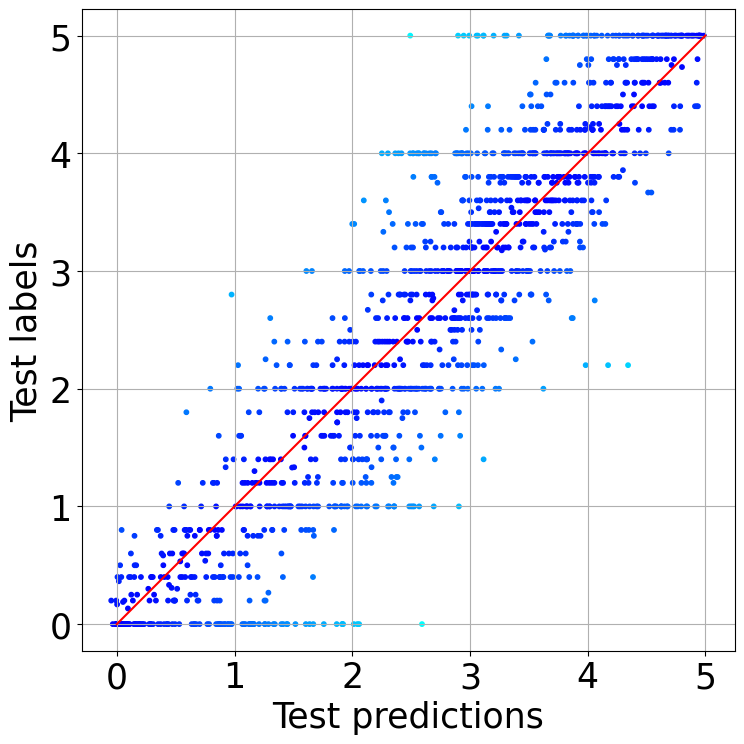}\par 
    \includegraphics[width=1\linewidth]{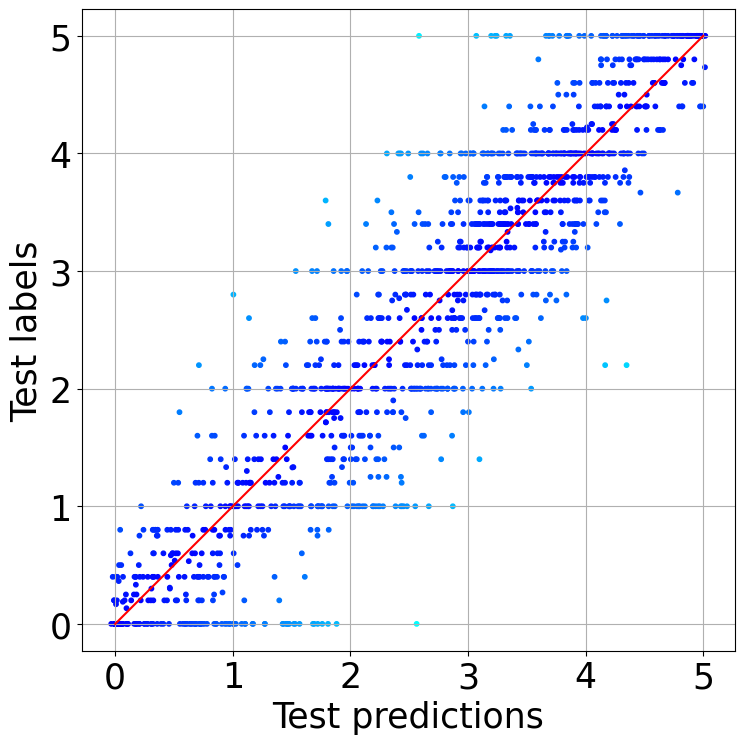}\par     
\end{multicols}
\end{center}
\end{figure}


\section{Results}
The baselines chosen are simple machine learning algorithms combined with averaged \emph{word2vec} token representation. The results are in table \ref{baseline-results}. All the baselines perform poorly, indicating that non-contextual token representations and inductive biases of these algorithms aren't powerful enough to determine sentence similarity. The best baseline approach is the unsupervised approach where the cosine similarity between two averaged \emph{word2vec} token representations is calculated.\\

The tables \ref{mse-results} and \ref{ce-results} illustrate the performance of transformer architectures on the STSB dataset. We can see that the transformer models outperform the baselines. Another important observation is that partial fine-tuning achieves decent results in most cases. The worst-performing models are the RoBERTa, distilBERT, and distilRoBERTa. We speculate this subpar perfomance is due to the representations not being good enough. RoBERTa does not use next sentence prediction, and distilled models have fewer parameters, so they do not understand language as BERT does. Interestingly, using either loss function approach to partial fine-tuning, DeBERTaV3 base achieves the best results even though it has only 42\% of the parameters of DeBERTaV3 large.\\

As expected, the results for end-to-end tuning are better than partial fine-tuning. Here, DeBERTaV3 large achieves the best results using either loss function. Also, we can observe that RoBERTa, distilBERT, and distilRoBERTa achieved comparable results, despite performing much worse when partially fine-tuning.\\

Finally, we apply the boosting algorithms whose performance can be seen in tables \ref{boost-mse-results} and \ref{boost-ce-results}. LightGBM and XGBoost give slightly better results in contrast to the AdaBoost algorithm, while LightGBM outperforms XGBoost in most cases by a small margin. An important observation regarding the choice of transformers is that the hyperparameter optimization procedure always selected a combination of different transformer architectures instead of combining smaller and larger models of the same architecture. We speculate this is due to different architectures excelling on different examples in the dataset.

\section{Discussion}

\subsection{Examining the dataset splits}

After training the ensembles we noticed that we achieved better development set results but worse test set results. For comparison, the best-performing ensemble model achieves worse results than the best single transformer model. To examine this better, we compared the label distributions in different splits using the Kolmogorov-Smirnov test. Table \ref{kolmogorov} shows the achieved results. At a significance level of 0.01, we reject the null hypothesis that the samples are drawn from the same distributions in the train and development set. The same is true for those drawn from the development and test set. At a significance level of 0.01, we cannot reject the null hypothesis regarding the samples from train and test sets. This result indicates that the problem might lie in the label distributions, which differ across the given splits.
To confirm, we retrain the best ensemble model and the corresponding transformers architectures using a stratified cross-validation split achieved by binning. For simplicity, we restrict ourselves only to retraining BERT large, DeBERTaV3 large, and RoBERTa large. The results in tables \ref{stratified-ce-results} and \ref{boost-stratified-ce-results} confirm our suspicions. We can see that the differences between the development and test correlation coefficients of the best-performing model decrease from 0.012/0.013 to 0.004/0.007.

\setlength\extrarowheight{1pt}
\begin{table}[t]
\begin{center}
\caption{Results of the Kolmogorov-Smirnov test. The double asterisk (**) denotes significance at a level of 0.01, and a single asterisk (*) significance at a level of 0.05}
\begin{tabular}{ccccc}
\textbf{ } & \textbf{Train and dev} & \textbf{Dev and test} & \textbf{Train and test} \\
\hline
\textbf{Statistic value} & 0.10512 & 0.08657 & 0.04509 \\
\textbf{p-value} & \textbf{6.680e-12**} & \textbf{3.854e-5**} & \textbf{2.095e-2*}
\end{tabular}
\label{kolmogorov}
\end{center}
\end{table}

\let\thefootnote\relax\footnotetext{All of the code is available here: \urlstyle{tt}\url{https://github.com/ir2718/semantic-similarity-scoring}}

\subsection{The edge of prediction range}
When visualizing the data we noticed that our ensemble models have issues with prediction close to the edge of the prediction range. This problem can be seen in the scatterplots in figure \ref{fig:scatterplots}. To tackle this problem, we will only observe examples labeled with 0 or 5. We noticed several sentences with high errors have more filler words than others, which inspired us to calculate the Jaccard index of lemmas for paired sentences and to repeat the same calculation without stopwords and for each of the sentences in a pair. Figure \ref{kdeplots} shows the densities of examples with an absolute error smaller than 1 (declared correct) and examples with an absolute error greater or equal to 1 (declared incorrect) approximated from these features. We can clearly see that some of these features can discriminate very well, i.e., the first column suggests that the density of the correctly classified examples has two modes that are very close to each other, while the density of the incorrectly classified examples has lower values in this area. The differences are even more prominent in the last two columns where the density of the correctly classified examples has a higher mean compared to the density of the incorrect ones. This insight suggests examples with a higher error on the edge of the prediction range have fewer meaningful lemmas.\\

\setlength\extrarowheight{2pt}
\setlength\tabcolsep{1.5pt}
\begin{table}[t]
\begin{center}
\caption{Mean of absolute error between predictions and labels in the given label span}
\begin{tabular}{ccccccc}
\textbf{Label span} & $(0, 0.5]$ & $(0.5, 1.5]$ &  $(1.5, 2.5]$ &  $(2.5, 3.5]$ &  $(3.5, 4.5]$ &  $(4.5, 5]$ \\
\hline
LGBM with MSE & 0.547 & 0.532 & 0.465 & 0.388 & 0.472 & 0.602 \\
LGBM with CE & 0.465 & 0.505 & 0.474 & 0.409 & 0.413 & 0.462\\

\end{tabular}
\label{label-span}
\end{center}
\end{table}

Table \ref{label-span} illustrates our models have issues with the edge of the prediction range. This occurrence is more noticeable in means of absolute error for the mean squared error ensemble, although still existent for the model that uses cross-entropy error.

\section{Conclusion}

In this paper, we explore combining modern transformer architectures such as BERT with traditional ensembling algorithms. Besides using outputs from 2 and 3 transformer architectures, we combine this information with handcrafted features. Our experiments suggest that this combination of algorithms achieves higher metric values on the development set while being detrimental to performance on the test set. We show that this might be due to the predefined dataset splits by repeating the experiments for our best-performing models using a stratified cross-validation version of the dataset. The results indicate that the models trained this way are more accurate on the development set and have better generalization properties on the test set. Finally, we provide an analysis of our best-performing ensemble models that suggests the number of meaningful lemmas is a good indicator of whether an example is difficult, and that these models are less accurate on the edges of the prediction range. Our work is a good starting point for other ideas regarding combining ensembling algorithms with transformers, such as investigating the effect of ensembling when using only small transformer models, extending this idea to sentence-embedding models such as SentenceBERT, as well as employing ensembling to specific textual domains.

\bibliographystyle{ieeetran}
\bibliography{bibliography.bib}
\nocite{*}
\end{document}